\documentclass[letterpaper]{article} % DO NOT CHANGE THIS
\usepackage{aaai2026}  % DO NOT CHANGE THIS
\usepackage{times}  % DO NOT CHANGE THIS
\usepackage{helvet}  % DO NOT CHANGE THIS
\usepackage{courier}  % DO NOT CHANGE THIS
\usepackage[hyphens]{url}  % DO NOT CHANGE THIS
\usepackage{graphicx} % DO NOT CHANGE THIS
\usepackage{natbib}  % DO NOT CHANGE THIS AND DO NOT ADD ANY OPTIONS TO IT
\usepackage{caption} % DO NOT CHANGE THIS AND DO NOT ADD ANY OPTIONS TO IT
\usepackage{algorithm}
\usepackage{algorithmic}

\usepackage{newfloat}
\usepackage{listings}
\DeclareCaptionStyle{ruled}{labelfont=normalfont,labelsep=colon,strut=off} % DO NOT CHANGE THIS
\floatstyle{ruled}
\newfloat{listing}{tb}{lst}{}
\floatname{listing}{Listing}
\usepackage{array}
\usepackage{amsmath}
\usepackage{graphicx}
\usepackage{xcolor}
\usepackage{subfig}
\usepackage{diagbox}
\usepackage{multirow}
\usepackage{algorithm}
\usepackage{listings}

\usepackage{amssymb}

\usepackage{times}
\usepackage{helvet}
\usepackage{courier}
\usepackage{xcolor}

\newcounter{checksubsection}
\newcounter{checkitem}[checksubsection]

\title{NTSFormer: A Self-Teaching Graph Transformer for Multimodal \\ Isolated Cold-Start Node Classification}
\author{
    Jun Hu, Yufei He, Yuan Li, Bryan Hooi, Bingsheng He
}
\affiliations{
    National University of Singapore\\
    jun.hu@nus.edu.sg, \{yufei.he, li.yuan\}@u.nus.edu, \{dcsbhk, dcsheb\}@nus.edu.sg
}

\usepackage{bibentry}
\begin{document}

\maketitle

\begin{abstract}

Isolated cold-start node classification on multimodal graphs is challenging because such nodes have no edges and often have missing modalities (e.g., absent text or image features). 
Existing methods address structural isolation by degrading graph learning models to multilayer perceptrons (MLPs) for isolated cold-start inference, using a teacher model (with graph access) to guide the MLP. 
However, this results in limited model capacity in the student, which is further challenged when modalities are missing.
In this paper, we propose \textbf{Neighbor-to-Self Graph Transformer (NTSFormer)}, a unified Graph Transformer framework that jointly tackles the isolation and missing-modality issues via a self-teaching paradigm. 
Specifically, NTSFormer uses a cold-start attention mask to simultaneously make two predictions for each node: a ``student'' prediction based only on self information (i.e., the node's own features), and a ``teacher'' prediction incorporating both self and neighbor information.
This enables the model to supervise itself without degrading to an MLP, thereby fully leveraging the Transformer’s capacity to handle missing modalities.
To handle diverse graph information and missing modalities, NTSFormer performs a one-time multimodal graph pre-computation that converts structural and feature data into token sequences, which are then processed by Mixture-of-Experts (MoE) Input Projection and Transformer layers for effective fusion.
Experiments on public datasets show that NTSFormer achieves superior performance for multimodal isolated cold-start node classification.

\end{abstract}


\begin{links}
    \link{Code}{https://github.com/CrawlScript/NTSFormer}
\end{links}

\section{Introduction}

Multimodal graphs, whose nodes are associated with diverse data modalities, are prevalent in real-world scenarios, such as social networks and product co-purchase networks.
\textbf{Node classification on multimodal graphs} is a critical task with applications in areas like fake news detection~\cite{zhang2024reinforced}, product tagging~\cite{zhu2024multimodalgraphbenchmark}, and more~\cite{DBLP:conf/aaai/CaiWLZ024}.
Multimodal Graph Neural Networks (GNNs) emerge as a promising solution by jointly modeling multimodal content and graph structure~\cite{DBLP:conf/mm/WeiWN0HC19,DBLP:journals/ipm/TaoWWHHC20}.

\begin{figure}[!t]
 % \vspace{-5mm}
    \centering
    \includegraphics[width=1.0\linewidth]{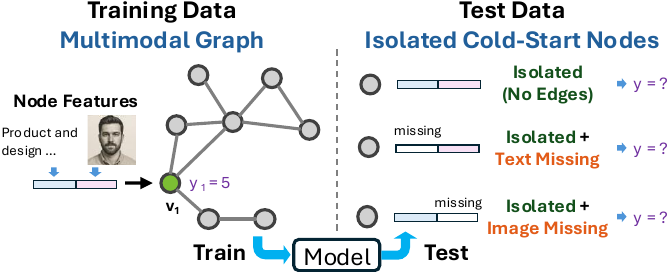}
    % \vspace{-2mm}
    \caption{
    The \textbf{multimodal isolated cold‑start node classification} task focuses on classifying isolated cold‑start nodes that have no edges and may be missing certain modalities.
    }
    \label{fig:intro_cold_start}
\vspace{-2mm}
\end{figure}

\textbf{Multimodal Isolated Cold-Start Node Classification.}
Real-world multimodal graphs frequently contain isolated cold-start nodes~\cite{moscati2024multimodal,shen2024enhancing}—newly introduced isolated nodes with limited information, such as newly registered users in social networks.
These isolated cold-start nodes typically pose two challenges for multimodal node classification models: (1) isolation, where the nodes have no connections~\cite{DBLP:conf/iclr/ZhangLSS22,DBLP:conf/icdm/HeM22,wang2024simmlptrainingmlpsgraphs}, and (2) missing modalities, where certain data (e.g., text or images) are absent~\cite{DBLP:conf/iclr/WuDTAN0F24,DBLP:conf/recsys/GanhorMHNS24}.
For example, as shown on the right side of Figure~\ref{fig:intro_cold_start}, a new user in a social network may have no connections (isolation) and only a profile image without providing a description (missing text).
Such scenarios make node classification particularly difficult, as models cannot leverage the graph structure and must instead depend on multimodal features, which are sometimes missing.

\begin{figure}[!tp]
% \vspace{-6mm}
    \centering
    \includegraphics[width=0.78\linewidth]{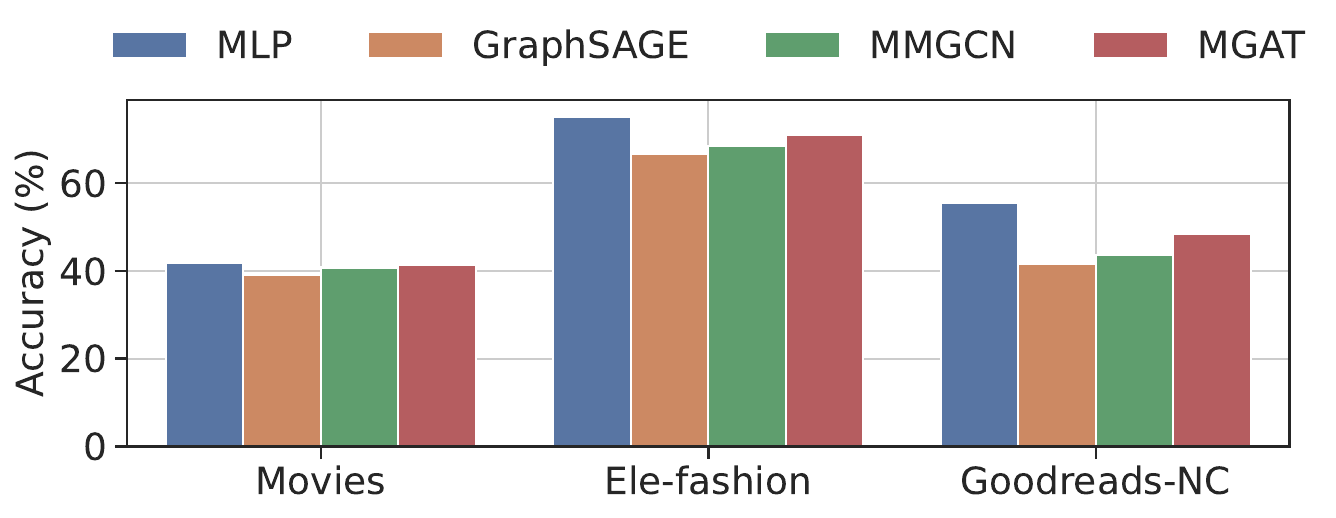}
% \vspace{-3mm}
\caption{
Performance on multimodal isolated cold-start node classification. 
General GNNs (GraphSAGE) and multimodal GNNs (MMGCN, MGAT) even underperform  MLPs.
}
    \label{fig:intro_mlp_and_gnns}
    \vspace{-4mm}
\end{figure}

To illustrate this challenge, Figure~\ref{fig:intro_mlp_and_gnns} presents the performance of multilayer perceptrons (MLPs) and GNNs on this task across public datasets.
Popular GNNs such as GraphSAGE~\cite{DBLP:conf/nips/HamiltonYL17}, MMGCN~\cite{DBLP:conf/mm/WeiWN0HC19}, and MGAT~\cite{DBLP:journals/ipm/TaoWWHHC20} generally outperform MLPs on non-cold-start tasks.
However, as illustrated in Figure~\ref{fig:intro_mlp_and_gnns}, these GNNs exhibit poor performance in isolated cold-start scenarios and can even be outperformed by simple MLPs.
The significantly degraded performance of GNN-based models on multimodal isolated cold-start node classification highlights the pressing need for robust strategies specifically designed for this task.

\begin{figure}[!tp]
\centering
\subfloat[
GLNN (MLP-Student)
]{
\label{fig:intro_mlp_stu}
\includegraphics[width=0.57\linewidth]{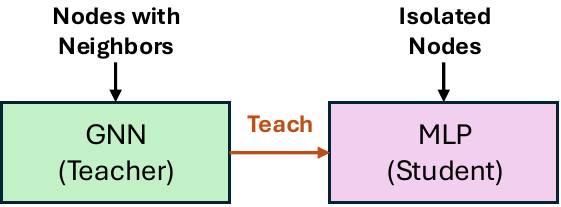}
}
\subfloat[
\label{fig:intro_self_teach}
Ours (Self-Teaching)
]{
\includegraphics[width=0.43\linewidth]{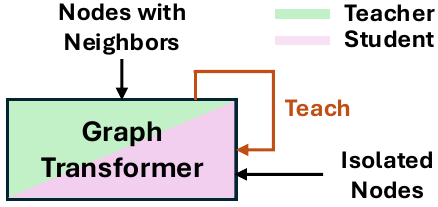}
}
\vspace{-1mm}
\caption{
NTSFormer uses a cold-start mask to make two predictions: a ``student'' prediction based on self-features only, and a ``teacher'' prediction with both self and neighbor information, enabling it to supervise itself without degrading to an MLP, thereby leveraging Transformers' capacity.
}
\label{fig:intro_mlp_stu_and_self_teach} 
\vspace{-3mm}
\end{figure}

\textbf{Teaching MLP-Students to Alleviate the Isolation Challenge.}
As shown in Figure~\ref{fig:intro_mlp_and_gnns}, MLP often outperforms GNNs in isolated cold-start scenarios.
One possible reason is that MLPs are trained and tested under consistent conditions, using only node-level features, whereas GNNs suffer from a train-test distribution shift due to the isolation of test-time nodes.
Despite this advantage, a major limitation of MLP-students is their inability to exploit graph structure during training, missing out on valuable relational signals.
To address this, recent methods such as GLNN~\cite{DBLP:conf/iclr/ZhangLSS22}, SGKD~\cite{DBLP:conf/icdm/HeM22}, and SimMLP~\cite{wang2024simmlptrainingmlpsgraphs} adopt a teacher-student paradigm, as illustrated in Figure~\ref{fig:intro_mlp_stu}, where a GNN teacher distills its structural knowledge into a structure-agnostic MLP-student.
This approach effectively bridges the gap caused by isolation and improves generalization in isolated cold-start settings.
However, such MLP-student methods often suffer from capacity bottlenecks in multimodal scenarios where they must handle more complex challenges, including missing modalities.

\textbf{From MLP-Students to Self-Teaching Graph Transformers.}
Motivated by the capacity limitations of MLP-students, we propose \textbf{Neighbor-to-Self Graph Transformer (NTSFormer)}, a unified Graph Transformer with a \emph{self-teaching} paradigm as shown in Figure~\ref{fig:intro_self_teach}.
Rather than resorting to a simple MLP, we harness the power of Transformers to handle both structural isolation and missing modalities.
Specifically, it produces two predictions for each node: a \emph{student} prediction based only on self-features, and a \emph{teacher} prediction that also incorporates neighbor information.  
This design ensures that training on graphs aligns seamlessly with inference on isolated cold-start nodes—eliminating the need to degrade the model to an MLP for such cases.  
To handle diverse graph data and missing modalities, NTSFormer performs a one-time multimodal graph pre-computation, converting both structural and multimodal feature information into token sequences, which are then processed by Mixture-of-Experts (MoE) Input Projection and Transformer layers for effective fusion.  
Extensive experiments on public datasets demonstrate that NTSFormer consistently outperforms various baseline methods.

Our main contributions are as follows:
\begin{itemize}
\item We introduce \textbf{NTSFormer}, a Graph Transformer that unifies the modeling of structural isolation and missing modalities in multimodal isolated cold-start settings via self-teaching.
\item To enable self-teaching within a unified Graph Transformer, we propose a \emph{cold-start mask} that enables a Graph Transformer to produce two predictions per node: a student prediction based solely on self-features and a teacher prediction incorporating neighbor information.
\item To handle diverse graph information and missing modalities, we design Multimodal Graph Pre-computation that captures various features from multimodal graphs, along with MoE Input Projection for effectively processing the various features before feeding them into Transformers.
\item We conduct extensive experiments on public multimodal graph benchmarks, demonstrating the superior performance of NTSFormer over various baselines.
\end{itemize}

\section{Related Work}

\subsection{Multimodal Graph Neural Networks}
GNNs such as GCN~\cite{DBLP:conf/iclr/KipfW17}, GAT~\cite{DBLP:conf/iclr/VelickovicCCRLB18}, and GraphSAGE~\cite{DBLP:conf/nips/HamiltonYL17} have achieved strong results for node classification by leveraging both node features and graph structure via message passing. 
To handle multimodal graphs, MMGCN~\cite{DBLP:conf/mm/WeiWN0HC19} uses modality-specific GNNs for intra- and inter-modal modeling. 
MGAT~\cite{DBLP:journals/ipm/TaoWWHHC20} applies attention to weight modalities. 
MIG~\cite{hu2024modalitymiggt} applies GNNs with modality-independent receptive fields.

\subsection{Multimodal Isolated Cold-Start Node Classification}

This task involves both structural isolation and missing modalities. 
Under structural isolation, traditional GNNs suffer from a train-test distribution shift, leading to degraded performance.
To address this, GLNN~\cite{DBLP:conf/iclr/ZhangLSS22} distills knowledge from a GNN teacher to an MLP student.
SA-MLP~\cite{chen2024samlp} employs structure-aware MLPs with structure-mixing distillation.
SGKD~\cite{DBLP:conf/icdm/HeM22} transfers GNN logits to MLPs using feature propagation.
SimMLP~\cite{wang2024simmlptrainingmlpsgraphs} applies self-supervised alignment between GNN and MLP.

For missing modalities, neighbor-based imputation methods like NeighMean~\cite{DBLP:conf/cikm/MalitestaRPNM24} and FeatProp~\cite{malitesta2024dealingmissingmodalitiesmultimodal} are inapplicable under isolation. 
MUSE~\cite{DBLP:conf/iclr/WuDTAN0F24} treats modalities as pseudo neighbors and applies modality dropping.
GMD~\cite{DBLP:conf/aaai/WangLHZ24} removes conflicting gradients to reduce modality co-dependence.
SiBraR~\cite{DBLP:conf/recsys/GanhorMHNS24} enforces modality consistency between shared encoders.

\subsection{Graph Transformers}

Graph Transformers (GTs) adapt Transformers for graphs and exhibit strong performance for node classification.
SGFormer~\cite{DBLP:conf/nips/WuZYZNJBY23} and Polynormer~\cite{DBLP:conf/iclr/DengYZ24} both use GNNs for local and Transformers for global encoding, but differ in integration.
NAGphormer~\cite{DBLP:conf/iclr/ChenGL023} precomputes neighbor features as tokens for Transformers on single-modality graphs, enabling classification without online message passing.

Different from existing work, we address multimodal isolated cold-start node classification by enabling self-teaching within Graph Transformers, allowing the model to simultaneously handle structural isolation and missing modalities.

\section{Problem Definition}

We study isolated cold-start node classification on multimodal graphs.
Formally, let the training graph be $\mathcal{G} = (\mathcal{V}, \mathbf{X}^{(t)}, \mathbf{X}^{(v)}, A, \mathbf{Y})$, where $\mathcal{V}$ is the set of nodes, and $N = |\mathcal{V}|$. 
Each node has a text feature vector of dimension $d_t$ and a visual feature vector of dimension $d_v$, with corresponding feature matrices $\mathbf{X}^{(t)} \in \mathbb{R}^{N \times d_t}$ and $\mathbf{X}^{(v)} \in \mathbb{R}^{N \times d_v}$.
The graph structure is given by the adjacency matrix $\mathbf{A} \in \{0,1\}^{N \times N}$.
The label vector $\mathbf{Y} \in \mathbb{Z}^{N}$ assigns each node a label in $\{1, \dots, C\}$ or $-1$ if unlabeled.

At test time, we are given a set of isolated cold-start nodes $\mathcal{V}_{\mathrm{te}}$. 
For simplicity, we still use $\mathbf{X}^{(t)}$, $\mathbf{X}^{(v)}$, $\mathbf{Y}$ to denote the text and visual features, and labels (as defined for training data) for the test nodes without subscripts.
Our goal is to predict their labels $\mathbf{Y}$. However, some test nodes may suffer from missing modalities, meaning that the corresponding text or visual feature vector is absent.

\begin{figure*}[!t]
    \centering
    \includegraphics[width=0.92\linewidth]{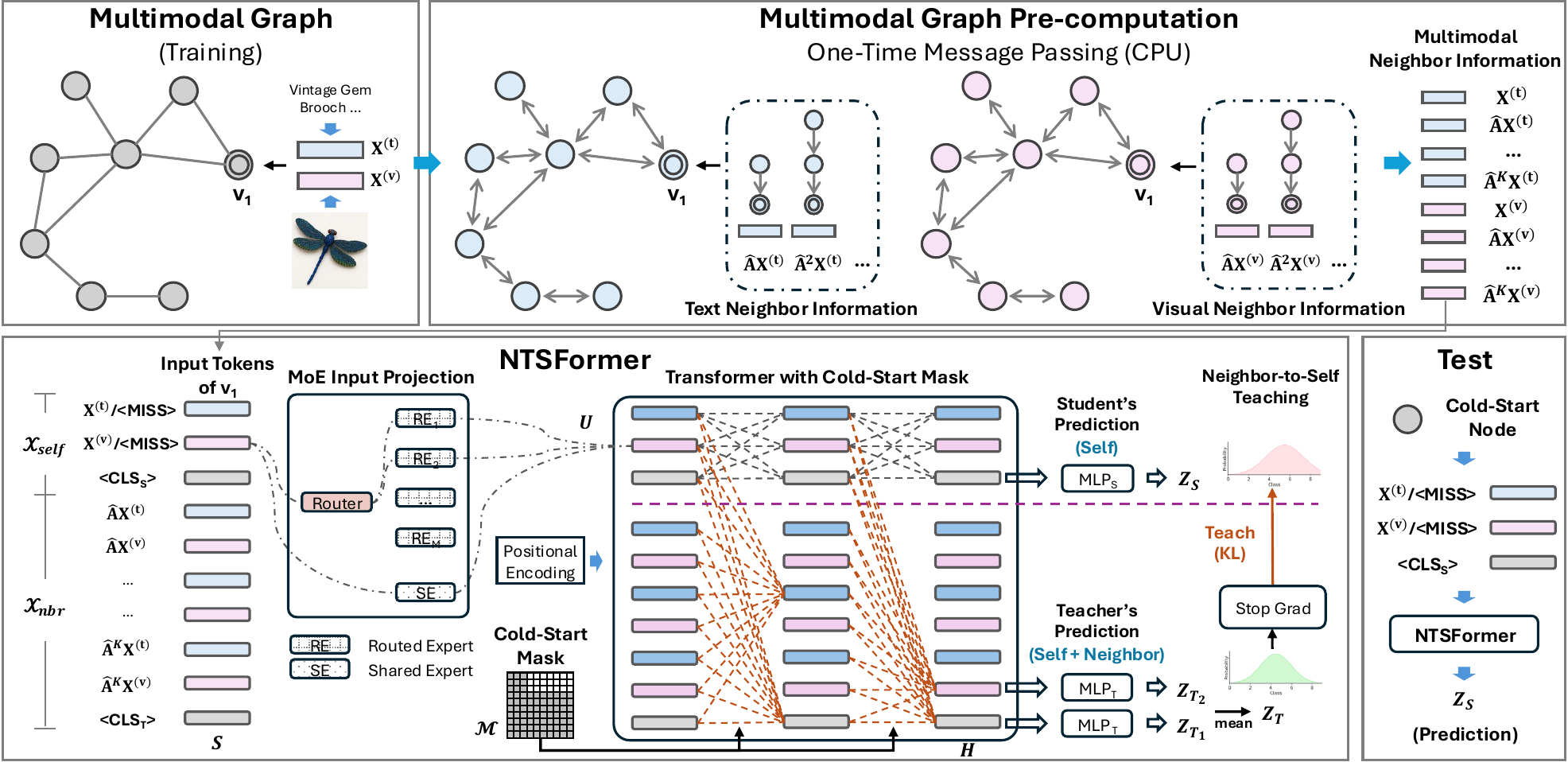}
\vspace{-2mm}
\caption{
Overall framework of NTSFormer.
}
 \label{fig:framework}
\vspace{-5mm}
\end{figure*}

\section{Method}

We present NTSFormer, a self-teaching Graph Transformer for multimodal isolated cold-start node classification.

\subsection{Overall Framework}

We propose \textbf{Neighbor-to-Self Graph Transformer (NTSFormer)}, a unified Graph Transformer framework specifically designed to address isolated cold-start node classification on multimodal graphs, handling both isolation (absence of edges) and modality missing (e.g., missing text or image features). 
It comprises three key modules (see Figure~\ref{fig:framework}):
\begin{itemize}
\item \textbf{Multimodal Graph Pre-computation}:
To capture diverse graph information for each node, this module performs a one-time pre-computation on multimodal graphs, converting multi-hop neighbor information across different modalities into fixed-length token sequences, which are then used as input to the Graph Transformer.

\item \textbf{Mixture-of-Experts (MoE) Input Projection}: To effectively fuse the diverse graph information, this module adopts multiple experts to dynamically project the input tokens into a shared embedding space, capturing different aspects of modality and neighbor information.

\item \textbf{Neighbor-to-Self Teaching via Cold-Start Masking}: To enable self-teaching of Graph Transformers for isolated cold-start node classification, this module applies a cold-start attention mask within a Transformer to simultaneously produce a \emph{student prediction} (from self-features only) and a \emph{teacher prediction} (with neighbor information), allowing the model to supervise itself without degrading to an MLP.
\end{itemize}

\subsection{Multimodal Graph Pre-computation}

To enable NTSFormer to capture rich information in multimodal graphs, we perform a one-time multimodal graph pre-computation~\cite{frasca2020sign} to convert neighbor information across different hops and modalities into token sequences suitable for Transformers.
Given node features—text $X^{(\mathrm{t})} \in \mathbb{R}^{N \times d_t}$ and visual $X^{(\mathrm{v})} \in \mathbb{R}^{N \times d_v}$—we first align feature dimensions by zero-padding each modality to $d_{in} = \max(d_t, d_v)$.
We then collect multimodal neighbor information within $K$ hops by computing $\{\hat{A}^k X^{(\mathrm{t})} | k = 1, \ldots, K\}$ for text modality and $\{\hat{A}^k X^{(\mathrm{v})} | k = 1, \ldots, K\}$ for visual modality, where $\hat{A} = \tilde{D}^{-1/2} \tilde{A} \tilde{D}^{-1/2}$ is the symmetrically normalized adjacency~\cite{DBLP:conf/iclr/KipfW17} with $\tilde{A} = A + I$ and $\tilde{D}_{ii} = \sum_{j=1}^{N} \tilde{A}_{ij}$. 
Here, $\hat{A}^k X^{(\mathrm{t})} \in \mathbb{R}^{N \times d_{in}}$ and $\hat{A}^k X^{(\mathrm{v})} \in \mathbb{R}^{N \times d_{in}}$ represent text and visual information collected from $k$-hop neighbors, respectively.

\textbf{Input Tokens Construction.}
We arrange the collected multimodal neighbor information into an ordered input token sequence matrix for NTSFormer.
We define the self-information tokens (self tokens) as:
\begin{equation}
\small
\mathcal{X}_{\text{self}} = [X^{(\mathrm{t})} \text{ or } \langle \mathrm{MISS} \rangle,\ X^{(\mathrm{v})} \text{ or } \langle \mathrm{MISS} \rangle,\ \langle \mathrm{CLS}_S \rangle]
\end{equation}
If a modality is missing at inference time, its position is replaced with a learned placeholder token $\langle \mathrm{MISS} \rangle$. 
During training, we simulate such cases by randomly replacing $X^{(\mathrm{t})}$ or $X^{(\mathrm{v})}$ with $\langle \mathrm{MISS} \rangle$ with probability $p_{\text{miss}}$. The $\langle \mathrm{CLS}_S \rangle$ token is a $d$-dimensional learnable vector used by the student for classification.
The neighbor tokens are defined as:
\begin{equation}
\small
\mathcal{X}_{\text{nbr}} = [\hat{A} X^{(\mathrm{t})},\ \hat{A} X^{(\mathrm{v})},\ \dots,\ \hat{A}^K X^{(\mathrm{t})},\ \hat{A}^K X^{(\mathrm{v})},\ \langle \mathrm{CLS}_T \rangle]
\end{equation}
where $\langle \mathrm{CLS}_T \rangle$ is another $d$-dimensional learnable token used by the teacher for classification.

The full input token sequence matrix is constructed as $S = \mathcal{X}_{\text{self}} \oplus \mathcal{X}_{\text{nbr}}$, where $\oplus$ denotes sequence-wise concatenation along the token dimension.
The length of input token sequence per node is $L = 2K + 4$, comprising $2(K + 1)$ modality tokens and two classification tokens.
In this paper, sequences are 1-indexed (e.g., $S[1]$ is the first item), and $S[-i]$ denotes the $i$-th item from the end.

This pre-computation is performed once on CPU, requires no gradient computation, and enables efficient Transformer training with fixed-length token sequences and standard mini-batching.

\subsection{MoE Input Projection}

Input projection layers are common in Transformers~\cite{DBLP:conf/iclr/DosovitskiyB0WZ21}, which map raw input feature tokens into a unified embedding space.
The typical choice—using a shared MLP uniformly for all tokens—however, is limited in our setting, since the input tokens come from various sources (e.g., self/neighbor, modality, or special tokens).

To address this, we design an input projection module based on MoE techniques~\cite{dai2024deepseekmoeultimateexpertspecialization,fedus2022switch} that improves the model's capacity to handle diverse data by dynamically routing each token to specialized expert networks. 
A gating network is used to determine how tokens are assigned to different experts, including multiple routed experts and one shared expert.

Our gating network considers token positions, since they  indicate tokens' semantic roles (e.g., self/neighbor, modality, or special token).
For the $i$-th token in the input sequence across all $N$ nodes, $S[i] \in \mathbb{R}^{N \times d_{in}}$, we concatenate a one-hot position vector $e_i \in \{0,1\}^{L}$ with only the $i$-th element equal to 1.
The combined input is $\tilde{S}[i] = [S[i] \,\|\, \mathbf{1}_N e_i^\top] \in \mathbb{R}^{N \times (d_{\text{in}} + L)}$, and the gating scores are computed as:
\begin{equation}
\small
\gamma = \mathrm{softmax}\left( \tilde{S}[i] \cdot W_{\text{gate}} \right) \in \mathbb{R}^{N \times M},
\end{equation}
where $W_{\text{gate}} \in \mathbb{R}^{(d_{\text{in}} + L) \times M}$ is a learnable weight matrix. 
The result $\gamma$ contains the normalized routing scores over the $M$ experts for each node at token position $i$.

Each routed expert is implemented as a 2-layer MLP, denoted $\mathrm{MLP}_{\text{RE}_m}: \mathbb{R}^{d_{\text{in}}} \rightarrow \mathbb{R}^{d}$, where $d$ is the hidden dimension of the Transformer. 
For each token position $i$, we select the top-$\hat{k}$ experts per node, indicated by a binary mask $\mathcal{T}(S[i]) = \mathrm{TopK}_{\text{row-wise}}(\gamma,\, \hat{k}) \in \{0,1\}^{N \times M}$, where $\mathcal{T}(S[i])_{j, m}=1$ indicates expert $m$ is selected for node $j$.
The routed expert output for token position $i$ and node $j$ is:
\begin{equation}
\small
S'_{\text{RE}}[i]_j = \sum_{m=1}^{M} \mathcal{T}(S[i])_{j, m} \cdot \gamma_{j, m} \cdot \mathrm{LN}\left( \mathrm{MLP}_{\text{RE}_m}(S[i]_j) \right),
\end{equation}
where $\gamma_{j, m} \in \mathbb{R}$ is the gating score, and $\mathrm{LN}$ is layer normalization. 
The full output matrix $S'_{\text{RE}}[i] \in \mathbb{R}^{N \times d}$ contains the routed representations for all $N$ nodes at token position $i$.

We also include a shared expert $\mathrm{MLP}_{\text{SE}}: \mathbb{R}^{d_{\text{in}}} \rightarrow \mathbb{R}^{d}$, and the final projected output for token position $i$ and node $j$ is:
\begin{equation}
\small
S'[i]_j = S'_{\text{RE}}[i]_j + \mathrm{LN}\left( \mathrm{MLP}_{\text{SE}}(S[i]_j) \right).
\end{equation}
Applying this projection across all token positions yields the MoE-projected sequence: $U = \mathrm{MoE}(S) \in \mathbb{R}^{N \times L \times d}$, where the projected tokens are combined into a 3D tensor over nodes, sequence length, and embedding dimension. 
This tensor serves as the input to the Transformer encoder.

\textbf{MoE Regularization.}
To promote balanced expert usage in the MoE input projection layer, we include a load balancing loss~\cite{fedus2022switch} $\mathcal{L}_{\mathrm{MoE}} = \sum_{m=1}^{M} P_m \cdot f_m$, where $P_m$ is the average gate score and $f_m$ is the fraction of tokens routed to expert $m$.

\subsection{Neighbor-to-Self Teaching via Cold-Start Masking}

To enable self-teaching within a unified Transformer, we design a mechanism that produces two distinct predictions:
\begin{itemize}
    \item The output token at \(\langle \mathrm{CLS}_S \rangle\) serves as the \textit{student} prediction, based solely on self information, and is directly compatible with isolated cold-start inference scenarios.
    \item The output token at \(\langle \mathrm{CLS}_T \rangle\) serves as the \textit{teacher} prediction, incorporating both self and neighbor information for supervision during training.
\end{itemize}
The teacher learns from full graph context, while the student mimics it using only self input. 

Using standard self-attention~\cite{vaswani2017attention} allows all tokens to attend to each other, causing neighbor information to leak into $\langle \mathrm{CLS}_S \rangle$ and breaking the cold-start isolation assumption.
To address this, we introduce a cold-start attention mask $\mathcal{M} \in \{0, 1\}^{L \times L}$ as follows to separate student and teacher contexts during attention:
\begin{equation}
\small
\mathcal{M} 
\hspace{-0.8mm}  = \hspace{-0.8mm} 
\begin{pmatrix}
\mathcal{M}^{(s \to s)} & \hspace{-2mm} \mathcal{M}^{(s \to n)} \\\
\mathcal{M}^{(n \to s)} & \hspace{-2mm} \mathcal{M}^{(n \to n)}
\end{pmatrix} 
\hspace{-0.8mm} = \hspace{-0.8mm}
\begin{pmatrix}
\mathbf{1}^{3 \times 3}     & \hspace{-2mm} \mathbf{0}^{3 \times (L-3)} \\\
\mathbf{1}^{(L-3) \times 3} & \hspace{-2mm} \mathbf{1}^{(L-3) \times (L-3)}
\end{pmatrix}
\hspace{-1.0mm}
\end{equation}
where $\mathcal{M}_{ij} = 0/1$ disables/allows attention from token $i$ to $j$.
Here, $3$ corresponds to the length of $\mathcal{X}_{\text{self}}$, which contains self information (including $\langle \mathrm{CLS}_S \rangle$) and occupies the first three positions of the input token sequence. 
Specifically, the block $\mathcal{M}^{(s \to n)}$ gates attention from \textbf{s}elf tokens to \textbf{n}eighbor tokens, which are placed after position $3$.
$\mathcal{M}$ therefore ensures that self tokens can only attend to each other and are strictly blocked from accessing neighbor tokens.
As a result, the representation at $\langle \mathrm{CLS}_S \rangle$ remains purely self-based and suitable for isolated cold-start prediction, while $\langle \mathrm{CLS}_T \rangle$ incorporates full context and serves as the teacher for self-supervision during training.

With the cold-start mask, we apply $L^{(\mathrm{tf})}$ self-attention Transformer layers after the MoE input projection.
Each layer updates the hidden states as follows:
\begin{equation}
\small
\begin{split}
H'^{(\ell)} &= \mathrm{LN}\left( \mathrm{MHA}(H^{(\ell-1)}; \mathcal{M}) + H^{(\ell-1)} \right)
\end{split}
\end{equation}
\begin{equation}
\small
\begin{split}
H^{(\ell)} &= \mathrm{LN}\left( \mathrm{FFN}(H'^{(\ell)}) + H'^{(\ell)} \right)
\end{split}
\end{equation}
where $H^{(1)} = U + \mathrm{PE}$ is the input to the first layer, and $\mathrm{PE} \in \mathbb{R}^{L \times d}$ is a learnable positional encoding.
The Multi-Head Attention (MHA) module computes attention as:
\begin{equation}
\small
\mathrm{MHA}(X; \mathcal{M}) = \mathrm{Concat}(\mathrm{head}_1, \dots, \mathrm{head}_h) W^{(O)},
\end{equation}
\begin{equation}
\small
\mathrm{head}_i = \mathrm{Att}\left(X W^{(Q)}_{(i)}, X W^{(K)}_{(i)}, X W^{(V)}_{(i)}\right),
\end{equation}
\begin{equation}
\small
\mathrm{Att}(Q, K, V) = \mathrm{softmax}\left( \frac{QK^\top}{\sqrt{d}} + (1 - \mathcal{M}) \cdot (-\infty) \right)V.
\end{equation}
Here, $W^{(Q)}_{(i)}, W^{(K)}_{(i)}, W^{(V)}_{(i)} \in \mathbb{R}^{d \times d_h}$ and $W^{(O)} \in \mathbb{R}^{hd_h \times d}$ are parameters.
The Feed-Forward Network (FFN) is a 2-layer MLP with GELU activation, applied independently to each token position with input/output dimension $d$.

\textbf{Prediction and Self-Teaching.}  
The output of the last Transformer, $H \in \mathbb{R}^{N \times L \times d}$, contains representations for all $L$ tokens.
The student prediction is produced from $\langle \mathrm{CLS}_S \rangle$ (token index: 3) via a student classifier $\mathrm{MLP}_S: \mathbb{R}^d \rightarrow \mathbb{R}^C$:
\begin{equation}
\small
Z_S = \mathrm{softmax}(\mathrm{MLP}_S(H[:, 3])), \quad Z_S \in \mathbb{R}^{N \times C}.
\end{equation}
Similarly, we use a teacher classifier $\mathrm{MLP}_T: \mathbb{R}^d \rightarrow \mathbb{R}^C$ for teacher predictions.
By default, we use the last token $\langle \mathrm{CLS}_T \rangle$ (index: -1):
\begin{equation}
\small
Z_{T_1} = \mathrm{softmax}(\mathrm{MLP}_T(H[:, -1])), \quad Z_{T_1} \in \mathbb{R}^{N \times C}.
\end{equation}
To provide more stable supervision, we follow prior practice~\cite{DBLP:conf/nips/BerthelotCGPOR19} and incorporate an additional teacher signal. 
Specifically, we take the second-to-last token (index: -2), which also attends to neighbor information, and compute a second prediction:
\begin{equation}
\small
Z_{T_2} = \mathrm{softmax}(\mathrm{MLP}_T(H[:, -2])), \quad Z_{T_2} \in \mathbb{R}^{N \times C}.
\end{equation}
The final teacher output is $Z_T = \frac{1}{2}(Z_{T_1} + Z_{T_2}) \in \mathbb{R}^{N \times C}$.

The student output $Z_S$ is trained to match the gradient-stopped teacher output $\mathrm{stopgrad}(Z_T)$ using a self-teaching loss $\mathcal{L}_{\mathrm{ST}}$ based on the Kullback–Leibler (KL) divergence:
\begin{equation}
\mathcal{L}_{\mathrm{ST}} = \mathrm{KL}(\mathrm{stopgrad}(Z_T) \| Z_S).
\end{equation}

\subsection{Optimization and Inference}

\textbf{Training Objective.}
We use the standard cross-entropy loss for teacher prediction: $\mathcal{L}_{\mathrm{CE}} = \mathrm{CrossEntropy}(Y, Z_T)$.
This is combined with the self-training loss and MoE regularization to form the final training objective:
\begin{equation}
\mathcal{L} = \mathcal{L}_{\mathrm{CE}} + \lambda \mathcal{L}_{\mathrm{ST}} + \gamma \mathcal{L}_{\mathrm{MoE}},
\end{equation}
where $\lambda$ and $\gamma$ is a tunable hyperparameter. The entire model is optimized end-to-end using the AdamW optimizer.

\textbf{Isolated Cold-Start Inference.}
At test time, NTSFormer takes in the self-only input sequence $\mathcal{X}_{\text{self}}$, where any missing modality is replaced with $\langle \mathrm{MISS} \rangle$. We take the student prediction from the output corresponding to $\langle \mathrm{CLS}_S \rangle$:
\begin{equation}
\small
Z_S = \mathrm{softmax}(\mathrm{MLP}_S(\mathrm{NTSFormer}(\mathcal{X}_{\text{self}})[:, 3])).
\end{equation}

\subsection{Complexity Analysis}
We analyze the time complexity of NTSFormer. 
For pre-computation, the complexity is $O(KEd_{in})$, where $E$ is the number of edges. 
Note that pre-computation is only required for training, not for inference.
For the model applied to each node, the complexities of MoE Input Projection, Multi-Head Attention, Feed-Forward Network, and prediction head are $O(KM(d_{in}d+d^2))$, $O(K^2d+Kd^2)$, $O(Kd^2)$, and $O(d^2+dC)$, respectively.

\section{Experiments}

We conduct experiments on public datasets.
All experiments are performed on a Linux system using a single NVIDIA RTX 3090 GPU (24GB), an Intel(R) Xeon(R) Gold 6226R CPU @ 2.90GHz, and 376 GB of RAM.

\begin{table}[!tp]
\centering
\scalebox{0.9}{

\small
\begin{tabular}{l c c c}\hline

Dataset      & Nodes  & Edges       & Classes   \\\hline
Movies       & 16,672    & 218,390     & 20 \\ \hline
Ele-fashion  & 97,766    & 199,602     & 12 \\ \hline
Goodreads-NC & 685,294   & 7,235,084   & 11 \\ \hline

\end{tabular}
}
\vspace{-1mm}
\caption{Statistics of datasets.}
\label{tab:datasets_statistics}
\vspace{-5mm}
\end{table}

\begin{table*}[!tp]
  \centering
\setlength{\tabcolsep}{0.8mm} 
\scalebox{0.8}{
\small
\begin{tabular}{l |c c c c | c c c c | c c c c}\hline
                     & \multicolumn{4}{ c |}{Movies}      & \multicolumn{4}{c|}{Ele-fashion}        & \multicolumn{4}{c}{Goodreads-NC}    \\\hline
               & Text-Miss & Visual-Miss & No-Miss & All & Text-Miss & Visual-Miss & No-Miss & All & Text-Miss & Visual-Miss & No-Miss & All \\\hline
MLP            & 41.15$\pm$1.33 & 37.63$\pm$1.23 & 46.76$\pm$2.60 & 41.85$\pm$1.34 & 69.72$\pm$3.61 & 71.67$\pm$1.51 & 84.04$\pm$0.51 & 75.15$\pm$0.71 & 40.25$\pm$0.83 & 58.16$\pm$1.01 & 68.27$\pm$0.54 & 55.56$\pm$0.22  \\\hline
GraphSAGE      & 38.60$\pm$3.04 & 37.05$\pm$1.41 & 41.40$\pm$2.46 & 39.02$\pm$2.02 & 74.68$\pm$1.23 & 46.57$\pm$0.47 & 78.42$\pm$0.41 & 66.56$\pm$0.23 & 36.25$\pm$2.45 & 34.47$\pm$3.35 & 54.23$\pm$0.92 & 41.65$\pm$1.01 \\
MMGCN          & 40.68$\pm$2.66 & 34.82$\pm$2.46 & 46.80$\pm$1.64 & 40.77$\pm$0.41 & 69.01$\pm$5.20 & 56.50$\pm$4.06 & 80.05$\pm$2.27 & 68.52$\pm$0.87 & 22.55$\pm$7.97 & 49.59$\pm$1.02 & 58.92$\pm$0.91 & 43.69$\pm$3.14 \\
MGAT           &  39.93$\pm$0.94 & 36.94$\pm$2.91 & 47.01$\pm$1.69 & 41.30$\pm$1.12 & 65.68$\pm$4.26 & 69.74$\pm$1.43 & 77.12$\pm$4.54 & 70.85$\pm$1.09 & 35.55$\pm$1.55 & 50.84$\pm$2.91 & 58.94$\pm$0.66 & 48.44$\pm$1.33 \\
MIG            &  39.57$\pm$2.53 & 36.91$\pm$2.72 & 46.04$\pm$2.64 & 40.84$\pm$1.32 & 65.94$\pm$2.82 & 70.19$\pm$5.07 & 81.90$\pm$1.66 & 72.67$\pm$1.85 &  37.65$\pm$2.64 & 46.35$\pm$5.11 & 59.51$\pm$1.84 & 47.84$\pm$2.37 \\
Polynormer     & 36.47$\pm$1.68 & 36.55$\pm$2.63 & 46.19$\pm$2.24 & 39.74$\pm$1.81 & 77.08$\pm$0.74 & 48.29$\pm$1.64 & 78.29$\pm$0.69 & 67.89$\pm$0.76 & 34.73$\pm$5.96 & 31.37$\pm$6.81 & 50.25$\pm$4.53 & 38.79$\pm$1.61  \\
SGFormer       & 41.23$\pm$1.48 & 39.28$\pm$3.10 & 46.22$\pm$2.06 & 42.25$\pm$0.60 & 77.62$\pm$0.88 & 49.69$\pm$1.46 & 79.94$\pm$0.74 & 69.08$\pm$0.60 & 33.21$\pm$2.73 & 36.13$\pm$1.60 & 56.20$\pm$1.49 & 41.85$\pm$1.07  \\
NAGphormer     & 39.03$\pm$2.25 & 39.14$\pm$1.39 & 44.50$\pm$2.52 & 40.89$\pm$1.48 & 71.08$\pm$0.99 & 70.90$\pm$2.58 & 82.41$\pm$0.36 & 74.79$\pm$0.87 & 35.74$\pm$1.00 & 46.66$\pm$0.97 & 56.95$\pm$0.57 & 46.45$\pm$0.53 \\
GLNN           & 41.26$\pm$1.58 & 39.25$\pm$1.95 & 48.60$\pm$1.18 & 43.04$\pm$0.85 & 67.11$\pm$3.47 & 71.53$\pm$6.25 & 84.60$\pm$0.55 & 74.41$\pm$1.53 & 35.88$\pm$1.23 & 57.44$\pm$0.60 & 68.71$\pm$0.48 & 54.01$\pm$0.33   \\
SGKD           & 42.81$\pm$1.68 & 37.59$\pm$1.65 & 46.76$\pm$1.27 & 42.39$\pm$0.77 & 72.73$\pm$2.61 & 73.23$\pm$2.76 & 84.33$\pm$0.51 & 76.76$\pm$1.09 & 40.19$\pm$0.89 & 58.26$\pm$0.63 & 68.50$\pm$0.54 & 55.65$\pm$0.38 \\ 
SimMLP         & 42.88$\pm$1.04 & 39.10$\pm$2.17 & 48.38$\pm$2.23 & 43.45$\pm$0.88 & 67.91$\pm$5.39 & 72.76$\pm$2.28 & 84.40$\pm$0.79 & 75.02$\pm$0.89 & 37.26$\pm$0.80 & 58.75$\pm$0.49 & 68.74$\pm$0.71 & 54.92$\pm$0.28  \\
SiBraR         & 36.55$\pm$1.90 & 40.51$\pm$1.94 & 44.17$\pm$1.73 & 40.41$\pm$0.79 & 70.51$\pm$3.90 & 47.19$\pm$5.54 & 73.27$\pm$3.78 & 63.65$\pm$2.97 & 28.59$\pm$5.17 & 31.55$\pm$2.84 & 34.70$\pm$4.16 & 31.62$\pm$0.92  \\
MUSE           & 42.34$\pm$1.75 & 39.82$\pm$1.40 & 48.16$\pm$1.56 & 43.44$\pm$1.13 & 77.43$\pm$0.84 & 80.53$\pm$0.50 & 84.01$\pm$0.33 & 80.66$\pm$0.34 & 34.48$\pm$1.00 & 51.67$\pm$1.15 & 59.34$\pm$0.75 & 48.49$\pm$0.59  \\

\hline

NTSFormer        & \textbf{45.07$\pm$1.12} & \textbf{42.52$\pm$1.41} & \textbf{50.79$\pm$1.89} & \textbf{46.12$\pm$0.58} & \textbf{80.71$\pm$0.89} & \textbf{83.98$\pm$0.80} & \textbf{85.42$\pm$0.54} & \textbf{83.37$\pm$0.46} & \textbf{50.99$\pm$0.13} & \textbf{63.93$\pm$0.36} & \textbf{69.83$\pm$0.47} & \textbf{61.58$\pm$0.16} \\\hline
               
\end{tabular}
}
% \vspace{-1mm}
\caption{Performance Comparison on Multimodal Isolated Cold-Start Node Classification (Accuracy (\%)).}
\label{tab:performance}
\vspace{-2mm}
\end{table*}

\subsection{Datasets}

We evaluate our approach on three real-world multimodal graph datasets from two benchmarks: \textbf{Movies} from the MAGB benchmark~\cite{yan2025graphmeetsmultimodalbenchmarking}, and \textbf{Ele-fashion} and \textbf{Goodreads-NC} (abbreviated as \textbf{GR-NC})  from the MM-Graph benchmark~\cite{zhu2024multimodalgraphbenchmark}, as shown in Table~\ref{tab:datasets_statistics}.

\textbf{Movies} is an Amazon movie-product e-commerce graph. Nodes are items with title/description text and cover images; edges indicate co-click or co-purchase interactions; labels are Amazon categories.
Text features use RoBERTa~\cite{liu2019robertarobustlyoptimizedbert}; image features use CLIP-ViT~\cite{DBLP:conf/icml/RadfordKHRGASAM21}.
\textbf{Ele-fashion} is an Amazon fashion-product graph. Nodes contain item-title text and product images; edges capture co-purchase relations; labels are fashion product categories (e.g., Shoes, Jewelry).
Text and image features are extracted with T5-Base~\cite{DBLP:journals/jmlr/RaffelSRLNMZLL20} and ViT-Base~\cite{DBLP:conf/iclr/DosovitskiyB0WZ21}, respectively.
\textbf{Goodreads-NC} is a large-scale graph derived from the Goodreads reading platform. 
Nodes are books with description text and cover images; edges link books with similar user preferences; labels are book genres (e.g., History, Children/Comics). 
Text and image features are extracted with T5-Base~\cite{DBLP:journals/jmlr/RaffelSRLNMZLL20} and ViT-Base~\cite{DBLP:conf/iclr/DosovitskiyB0WZ21}, respectively.

\subsection{Baselines}

We compare our method against a diverse set of baselines.
This includes MLP, which uses only node features without graph structure; a typical GNN, GraphSAGE~\cite{DBLP:conf/nips/HamiltonYL17}; multimodal GNNs, including MMGCN~\cite{DBLP:conf/mm/WeiWN0HC19}, MGAT~\cite{DBLP:journals/ipm/TaoWWHHC20}, and MIG~\cite{hu2024modalitymiggt}; GTs, including Polynormer~\cite{DBLP:conf/iclr/DengYZ24}, SGFormer~\cite{DBLP:conf/nips/WuZYZNJBY23}, and NAGphormer~\cite{DBLP:conf/iclr/ChenGL023}; models designed for isolated node classification, including GLNN~\cite{DBLP:conf/iclr/ZhangLSS22}, SGKD~\cite{DBLP:conf/icdm/HeM22}, and SimMLP~\cite{wang2024simmlptrainingmlpsgraphs}; and models for modality-missing scenarios, including SiBraR~\cite{DBLP:conf/recsys/GanhorMHNS24} and MUSE~\cite{DBLP:conf/iclr/WuDTAN0F24}.

\subsection{Experiment Settings and Evaluation Metrics}

\textbf{Dataset Partitioning.}
To model the isolated cold-start scenario, we partition each dataset into four subsets:  20\% labeled training nodes, 60\% unlabeled training nodes, 10\% validation nodes (isolated cold-start), and 10\% test nodes (isolated cold-start). 
The unlabeled training nodes are used in the training graph for message passing but receive no supervision. 
For validation and test nodes, all edges involving these nodes are removed to ensure they remain completely isolated, thereby creating an isolated cold-start environment.

\textbf{Missing-Modality Setting.}
We further evaluate performance on isolated cold-start nodes under a modality-missing scenario. 
Specifically, while all training nodes have complete modality information, the validation/test set is divided into three disjoint subsets of equal size: one missing text, one missing visual features, and one with full multimodal input.
That is, each subset contains approximately 33.3\% of the validation/test nodes.
We denote these three test conditions as \textbf{Text-Miss}, \textbf{Visual-Miss}, and \textbf{No-Miss}, and additionally report results on \textbf{All}, which combines all test nodes.

\textbf{Parameter Setting.}  
We apply unified hyperparameter settings across NTSFormer and all baselines to ensure fair comparison. 
In particular, the hidden dimension is fixed to 512, the maximum hop size is set to $K = 2$, and the input and hidden dropout rates are 0.2 and up to 0.5, respectively. 
All models are optimized using AdamW with learning rate $2 \times 10^{-3}$ and weight decay $1 \times 10^{-2}$. 
Training is run for 300 epochs on \textit{Movies} and \textit{Ele-fashion}, and 50 epochs on \textit{Goodreads-NC}. 
Early stopping is applied based on validation accuracy, and the best-performing checkpoint is used for final testing. Baselines are trained using their official implementations when available.
For NTSFormer, we employ additional method-specific settings. 
Specifically, NTSFormer uses $L^{(\mathrm{tf})} = 2$ Transformer layers with 2 attention heads, and the MoE input projection module comprising $M = 6$ routed experts along with 1 shared expert. 
The self-teaching and MoE regularization weights are $\lambda = 1.0$ and $\gamma = 0.1$, respectively.
These hyperparameters were selected based on a light random search over the validation set.

\textbf{Evaluation Metrics.}
We report classification accuracy as the evaluation metric.
All results are averaged over 5 runs with random seeds from 1 to 5, controlling for variation in both dataset splitting and model initialization. 
Importantly, with a fixed seed, all methods are evaluated on identical train/validation/test splits and identical modality-missing settings, ensuring fair comparison.
We report mean accuracy and standard deviation for each of the four settings: Text-Miss, Visual-Miss, No-Miss, and All.

\subsection{Performance Analysis}

Table~\ref{tab:performance} compares the performance of baselines and NTSFormer, from which we draw the following observations:
\begin{itemize}
\item General (multimodal) GNNs---GraphSAGE, MMGCN, MGAT, and MIG---do not perform well on this task and even underperform MLPs, verifying that isolated cold-start is a severe issue for GNNs on multimodal graphs.
\item Graph Transformers, including Polynormer, SGFormer, and NAGphormer, perform similarly to general GNNs and multimodal GNNs and do not perform well under the isolated cold-start setting.
\item GLNN, SGKD, and SimMLP are GNNs considering structural isolation, and outperform General (multimodal) GNNs and GTs.
Their MLP students taught by GNNs outperform the original MLP baseline.
\item SiBraR and MUSE are GNNs considering modality missing.
% %
MUSE shows superior performance over all general (multimodal) GNNs and GTs, and even over GNNs considering structural isolation on two datasets, showing that modality missing is also an important issue on this task.
SiBraR does not perform well, potentially due to its shared encoder design, which cannot handle each modality differently.
\item NTSFormer tackles isolation and modality-missing issues via a self-teaching paradigm, consistently outperforming all baselines across all datasets by a large margin, showing the advantage of self-teaching GTs on this task.
Compared to teacher-student baselines, we can directly perform self-teaching via end-to-end training, without requiring complicated two-step operations.

\end{itemize}

\begin{figure}[!t]
\centering
\subfloat[
Text-Miss
]{
\includegraphics[width=0.5\linewidth]{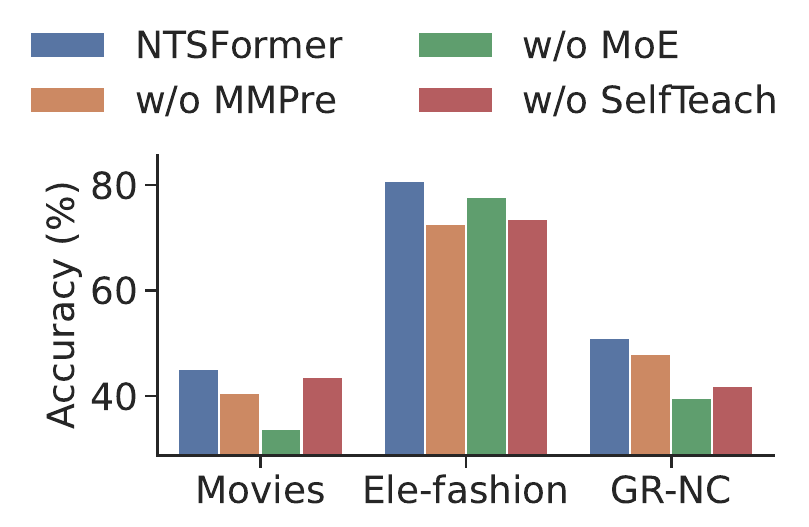}
}
\subfloat[
Visual-Miss
]{
\includegraphics[width=0.5\linewidth]{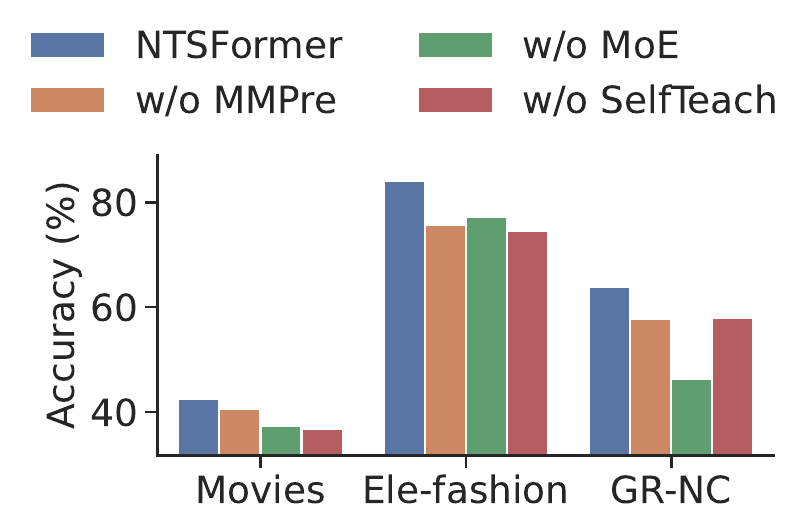}
}
\\
\subfloat[
No-Miss
]{
\includegraphics[width=0.5\linewidth]{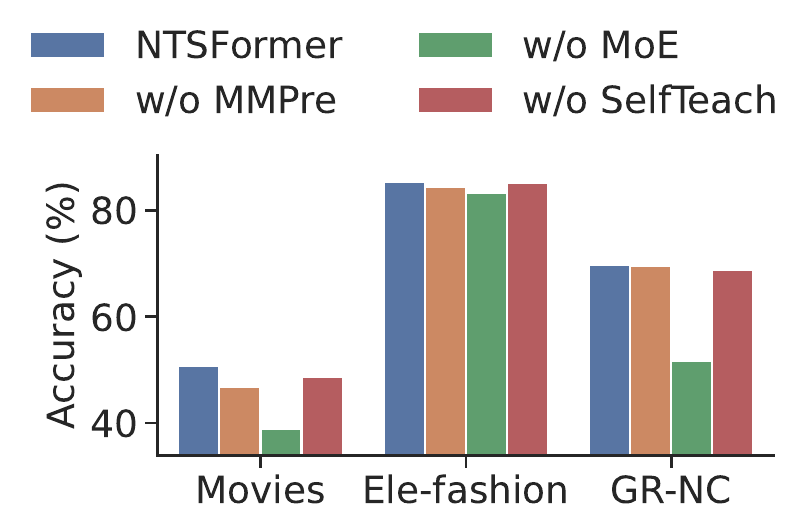}
}
\subfloat[
All
]{
\includegraphics[width=0.5\linewidth]{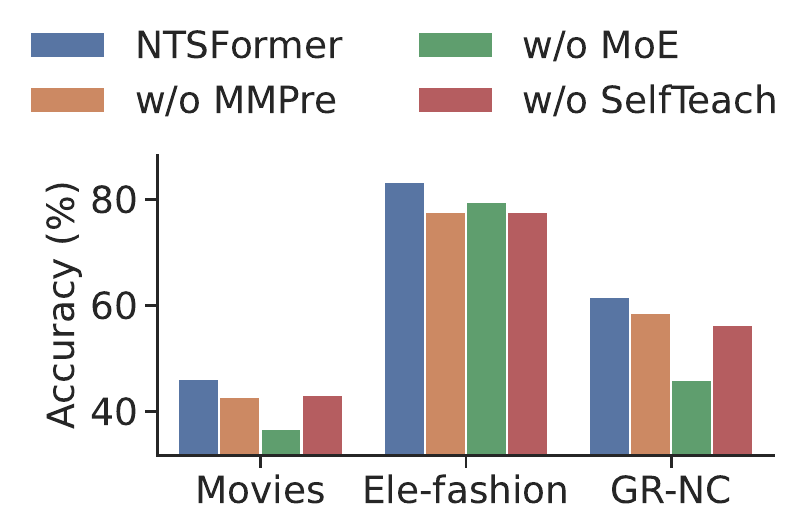}
}
\vspace{-2mm}
\caption{
Ablation results of \textbf{NTSFormer} and its ablated variants (w/o MMPre, w/o MoE, w/o SelfTeach).
The first three subfigures report performance on different subsets of test nodes, each with a specific modality-missing setting:
(a)~Text-Miss, 
(b)~Visual-Miss, 
and (c)~No-Miss. 
The last subfigure (d) reports performance on all test nodes.
}
\vspace{-4mm}
\label{fig:ablation_no_each} 
\end{figure}

\subsection{Detailed Analysis}

\subsubsection{Impact of Multimodal Graph Pre-computation}

Our pre-computation collects text and visual neighbor information into separate tokens. 
% %
The variant, NTSFormer w/o MMPre, replaces it with typical pre-computation.
Specifically, it concatenates text and visual features as node input, and the resulting tokens do not distinguish modalities.
Figure~\ref{fig:ablation_no_each} shows that NTSFormer consistently outperforms NTSFormer w/o MMPre, verifying our method's effectiveness.

\subsubsection{Impact of MoE Input Projection}

\begin{figure}[!t]
\centering
\subfloat[
Text-Miss
]{
\includegraphics[width=0.5\linewidth]{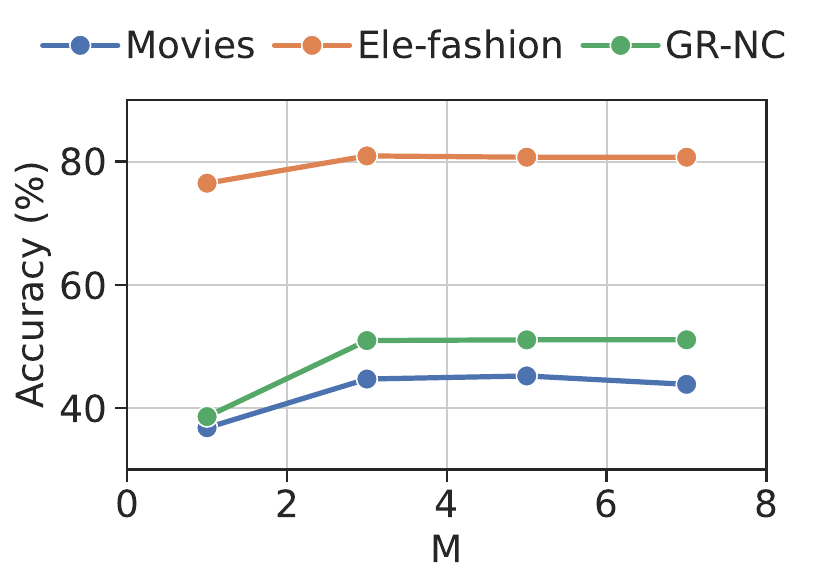}
}
\subfloat[
Visual-Miss
]{
\includegraphics[width=0.5\linewidth]{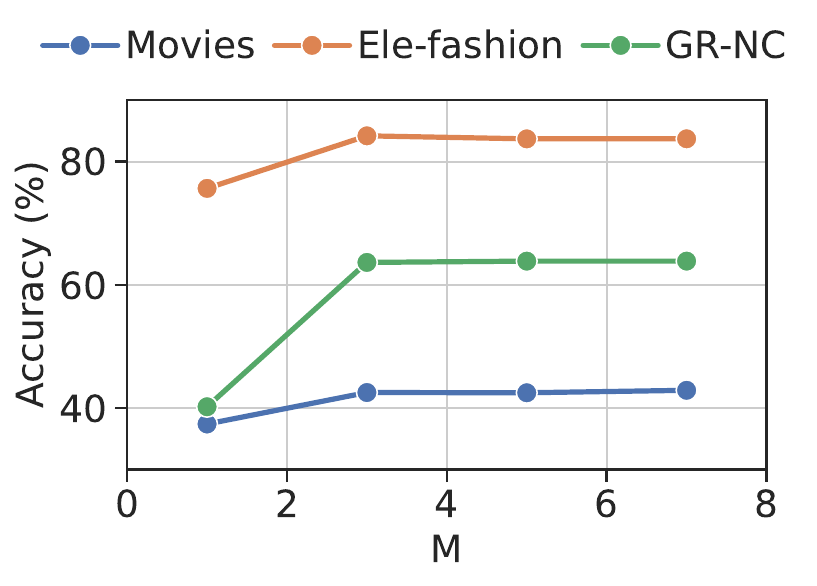}
}
\\
\subfloat[
No-Miss
]{
\includegraphics[width=0.5\linewidth]{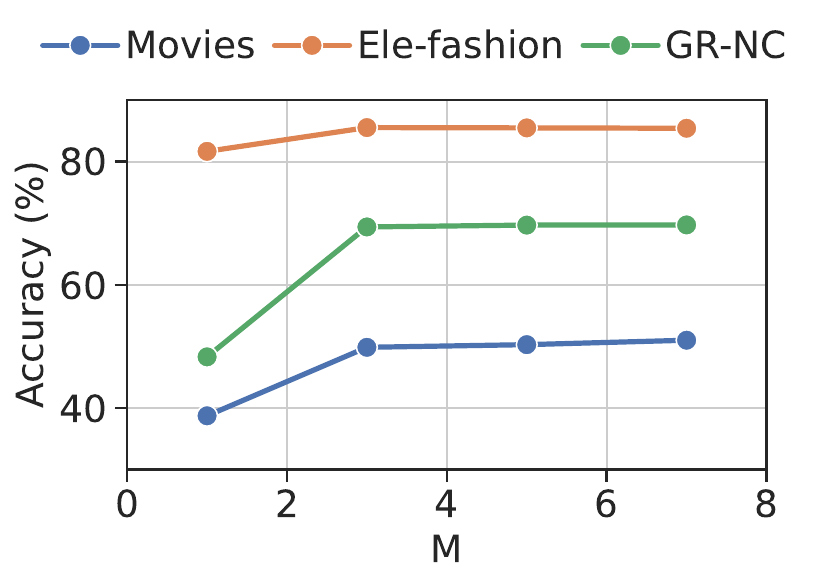}
}
\subfloat[
All
]{
\includegraphics[width=0.5\linewidth]{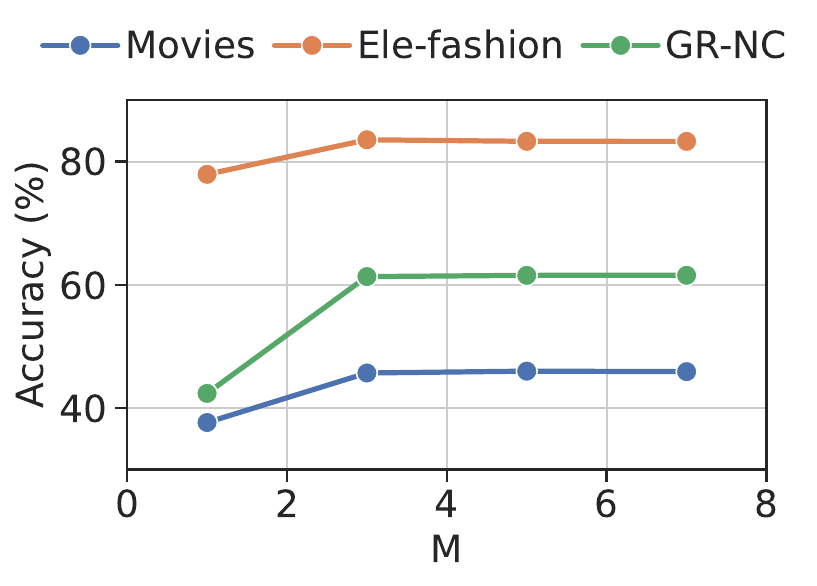}
}
\vspace{-2mm}
\caption{Impact of number of routed experts ($M$).}
\label{fig:impact_num_routed_experts} 
\end{figure}

To evaluate its effectiveness, we introduce a variant \textbf{NTSFormer w/o MoE}, where the MoE is replaced with a standard shared linear projection, following the design used in NAGphormer~\cite{DBLP:conf/iclr/ChenGL023}.
As shown in Figure~\ref{fig:ablation_no_each}, the variant consistently underperforms NTSFormer, verifying the effectiveness of MoE input projection.
Furthermore, Figure~\ref{fig:impact_num_routed_experts} illustrates the effect of varying the number of routed experts, $M$.
The results indicate that performance degrades when $M < 3$, underscoring the importance of employing multiple experts for effectively encoding diverse multimodal graph information.

\subsubsection{Impact of Self-Teaching Mechanism}

We replace the self-teaching mechanism with the MLP-student's two-branch setup, similar to GLNN~\cite{DBLP:conf/iclr/ZhangLSS22}.
In this variant, \textbf{NTSFormer w/o SelfTeach}, the student is implemented as a separate MLP trained independently to match the teacher branch output.
As shown in Figure~\ref{fig:ablation_no_each}, this setup leads to a noticeable drop in performance, showing the superiority of self-teaching GTs over the MLP-student method.

\subsubsection{Impact of Number of Transformer Layers}

As shown in Table~\ref{tab:impact_tf_layers}, using multiple layers generally provides a slight performance gain compared to a single layer. 
Although the best-performing layer depth varies per dataset—2 layers for Movies, 4 layers for Ele-fashion, and 3 layers for Goodreads-NC—none of the highest results occur at a single-layer configuration, suggesting that deeper architectures can offer marginal improvements.

\subsection{Efficiency of Training}

We analyze the training efficiency across datasets of varying sizes. 
Table~\ref{tab:training_efficiency} shows NTSFormer maintains comparable efficiency on smaller datasets (Movies and Ele-fashion) but exhibits advantages on the larger Goodreads-NC dataset. 
This improvement stems from our one-time pre-computation that converts graph structures into regular-shaped tensors, avoiding costly repetitive message passing during training~\cite{DBLP:journals/tkde/HuHH24}.
This enables better scalability on large graphs while achieving competitive performance.

\begin{table}[!tp]
  \centering
\scalebox{0.85}{
\small
\begin{tabular}{l c c c}
\hline
$L^{(\mathrm{tf})}$ & Movies & Ele-fashion & Goodreads-NC \\
\hline
1 & 45.22 & 83.54 & 61.36 \\
2 & \textbf{46.12} & 83.37 & 61.58\\
3 & 45.27 & 83.51 & \textbf{61.68} \\
4 & 45.32 & \textbf{83.67} & 61.51 \\
\hline
\end{tabular}
}
\vspace{-1mm}
\caption{Impact of number of Transformer layers ($L^{(\mathrm{tf})}$).}
\label{tab:impact_tf_layers}
\end{table}

\begin{table}[!tp]
  \centering
  
\scalebox{0.85}{
\small
\begin{tabular}{l c c c}
\hline
Method & Movies & Ele-fashion & Goodreads-NC \\
\hline
MIG & 82s & \textbf{96s} & 387s \\
GLNN & 65s & 161s & 778s \\
MUSE & 154s  &  406s & 1437s \\
NTSFormer & \textbf{46s} & 226s & \textbf{260s} \\
\hline
\end{tabular}
}
\vspace{-1mm}
\caption{Training efficiency comparison (seconds).}
\label{tab:training_efficiency}
\vspace{-4mm}
\end{table}

\section{Conclusion}

We study isolated cold-start node classification on multimodal graphs, where such nodes have no edges and often have missing modalities.
Existing methods degrade graph models to MLPs for cold-start inference, limiting capacity.
We propose the \textbf{Neighbor-to-Self Graph Transformer (NTSFormer)}, which addresses isolation and missing-modality issues via self-teaching without degrading to an MLP.
Experiments on public benchmarks show the superiority of NTSFormer over various baselines, underscoring the effectiveness of self-teaching Graph Transformers on this task.
Future work includes adapting it to dynamic graphs.

\section{Acknowledgments}

This research is supported by the National Research Foundation, Singapore and Infocomm Media Development Authority under its Trust Tech Funding Initiative, and the Ministry of Education, Singapore, under the Academic Research Fund Tier 2 (FY2025) (Grant MOE-T2EP20124-0009). Any opinions, findings and conclusions or recommendations expressed in this material are those of the author(s) and do not reflect the views of National Research Foundation, Singapore and Infocomm Media Development Authority.

\bibliography{aaai2026}

\end{document}